\documentclass[10pt,twocolumn,letterpaper]{article}

\usepackage[pagenumbers]{cvpr} 

\definecolor{cvprblue}{rgb}{0.21,0.49,0.74}
\usepackage[pagebackref,breaklinks,colorlinks,allcolors=cvprblue]{hyperref}

\usepackage{graphicx}
\usepackage{amsmath}
\usepackage{bm}
\usepackage{multirow}
\usepackage{bbding}
\usepackage{booktabs}
\usepackage{tabularx}
\usepackage{makecell}
\usepackage{microtype}

\def\paperID{*****} 
\def\confName{CVPR}
\def\confYear{2026}

\title{Multimodal LLM-Empowered Re-Ranking for Generalizable Person Re-Identification}

\author{Jiachen Li$^1$ and Xiaojin Gong\thanks{Corresponding author.}\\
College of Information Science and Electronic Engineering\\
Zhejiang University\\
{\tt\small $^1$12031134@zju.edu.cn, $^*$gongxj@zju.edu.cn}
}

\begin{document}
\maketitle

\begin{abstract}
    Domain Generalizable (DG) person re-identification (Re-ID) has attracted growing research interest due to its potential for deployment in unseen real-world scenarios. Most existing approaches address DG Re-ID by focusing on training domain-generalizable encoders but ignore the possible refinements in inference stage. In contrast, this work explores an alternative direction which improves inference re-ranking to enhance DG Re-ID. Conventional re-ranking methods typically rely on neighborhood-based distances to refine the initial ranking list, inherently depending on features produced by the Re-ID encoder. However, they deteriorate on target domains since the encoder lacks sufficient generalizability to produce reliable feature distances on unseen scenarios. Inspired by the remarkable generalization capabilities of recent Multimodal Large Language Models (MLLMs), we propose an MLLM-empowered distance metric to improve re-ranking in DG Re-ID. Specifically, we first adapt an MLLM to Re-ID data through supervised fine-tuning, which incorporates a domain-agnostic prompt and a query-candidate hard mining scheme. Then, the adapted MLLM is employed to compute a $\mu$-distance during inference, which is robust to domain gap and significantly enhances subsequent re-ranking performance. Our approach is model-agnostic and can be seamlessly integrated into previous re-ranking frameworks. Extensive experiments demonstrate that our approach consistently yields substantial performance improvements across multiple DG Re-ID benchmarks. The code of this work will be released at \href{https://github.com/RikoLi/MUSE}{https://github.com/RikoLi/MUSE} soon.
\end{abstract}

\section{Introduction}
\label{sec:introduction}
Person re-identification (Re-ID) employs encoders to extract ID-discriminative features for cross-camera matching. Despite significant progress, the generalization ability of Re-ID models across diverse domains still remains limited, hindering deployment in real-world scenarios. To address this issue, Domain Generalizable (DG) Re-ID, where models are trained on known source domains and evaluated on unseen target domains, has been studied to bridge domain gaps. Existing DG Re-ID approaches primarily focus on learning domain-invariant feature representations, either by designing more complex encoders~\cite{ADIN,MDA,META,QAConv,TransMatcher,PAT} or by leveraging more data~\cite{ISR,MMET,DMRL,ReMiX}.

In contrast to previous approaches, our work focuses on the re-ranking procedure to improve the performance of DG Re-ID. Re-ranking methods~\cite{CUHK03-NP,ECN,CAJ,CEIL} have been widely adopted as post-processing techniques in traditional intra-domain Re-ID tasks. Typically, an initial ranking list is generated based on Euclidean distance or cosine similarity computed from image features extracted by a Re-ID encoder, with top-ranked results considered as correct matches. However, due to the interference of ID-irrelevant factors such as illumination variations, viewpoint changes, and occlusions, some hard negative samples are often erroneously ranked high in the initial list. To mitigate such mis-rankings, re-ranking is applied to refine the retrieval results by leveraging the neighborhood context among gallery samples, thereby improving overall Re-ID accuracy.

\begin{figure*}[t]
	\centering
	\includegraphics[width=\textwidth]{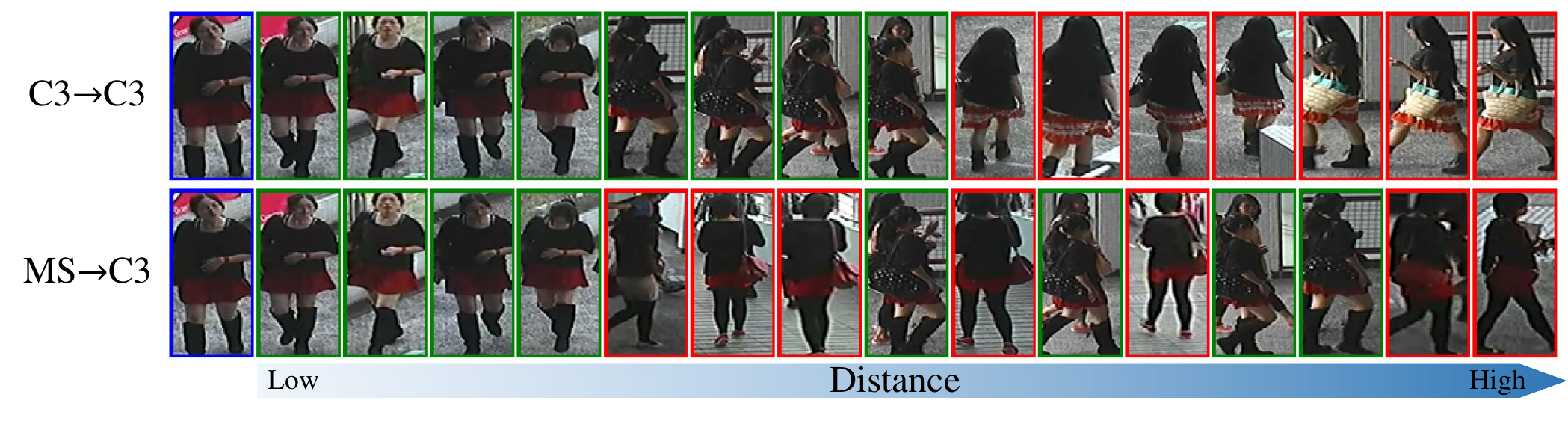}
	\caption{Illustration of re-ranking degradation in DG Re-ID. ``C3$\to$C3'' and ``MS$\to$C3'' indicate Re-ID encoders trained on source domains CUHK03-NP (C3)~\cite{CUHK03-NP} and MSMT17 (MS)~\cite{MSMT17}, respectively, and tested in target domain C3 with K-RNN re-ranking~\cite{CUHK03-NP}. The query image is highlighted with blue boxes. Top-15 closest retrievals after re-ranking are appended with ascending order of distance. Green and red boxes indicate true and false matching, respectively.}
	\label{fig:intra_vs_dg}
\end{figure*}

Typical re-ranking methods, such as K-RNN~\cite{CUHK03-NP}, utilize neighborhood-based metrics like Jaccard distance to refine the initial ranking list. Although effective in intra-domain settings, these re-ranking methods suffer from notable performance degradation when applied to unseen domains, as illustrated in Figure~\ref{fig:intra_vs_dg}. This limitation stems from their inherent reliance on features generated by Re-ID encoders trained on source domains. When these encoders lack sufficient domain generalizability, the extracted features fail to preserve consistent ID semantics across domains, leading to unreliable neighborhood structures and suboptimal re-ranking performance.

In this work, inspired by the remarkable generalization capabilities of Multimodal Large Language Models (MLLMs)~\cite{LLaVA, QwenVL,InternVL,DeepSeek-VL} exhibited in various language and vision tasks, we propose an \underline{Mu}ltimodal LLM-empowered Di\underline{s}tance M\underline{e}tric (MUSE) to incorporate MLLM's generalization power into DG Re-ID for improved re-ranking. Specifically, we first perform Supervised Fine-Tuning (SFT) to adapt an MLLM to Re-ID data, where the MLLM is fed with a query and a candidate image and asked to determine whether they belong to the same ID or not. To inject domain-generalizable knowledge during fine-tuning, we design a set of rules implemented via a domain-agnostic prompt that specifies the key attributes relevant for identity discrimination. To guarantee the efficiency and effectiveness of the tuning process, we introduce a Query-Candidate Hard Mining (QCHM) strategy to ensure that the model is only fine-tuned on challenging query-candidate pairs. During inference, we propose a novel $\mu$-distance\footnote{The symbol ``$\mu$'' denotes \underline{Mu}ltimodal LLM-empowered.} metric, which fuses the conventional Euclidean distance with an MLLM-predicted distance derived through pointwise likelihood generation~\cite{nogueira2020document}. This fusion yields a robust similarity measure resilient to domain shifts. Finally, our $\mu$-distance can be seamlessly integrated into existing re-ranking methods, consistently yielding substantial performance gains in DG Re-ID.

In summary, our main contributions are as follows:
\begin{itemize}
	\item To introduce the generalization capability of MLLMs into DG Re-ID for re-ranking enhancement, we adapt an MLLM through supervised fine-tuning, guided by a domain-agnostic prompt for injecting Re-ID knowledge and a query-candidate hard mining strategy for effective training.
	\item We propose $\mu$-distance, a robust distance metric for target domains, obtained by fusing the conventional Euclidean distance with an MLLM-predicted distance. This metric is compatible with various existing re-ranking methods.
	\item We conduct extensive experiments under multiple DG Re-ID evaluation protocols to thoroughly validate the effectiveness of our proposed approach, demonstrating that it significantly improves re-ranking performance and consistently outperforms existing state-of-the-art methods across various unseen target domains.
\end{itemize}

\section{Related Works}
\subsection{Generalizable Person Re-ID}
DG Re-ID has received considerable attention in recent years.  Classical approaches such as ADIN~\cite{ADIN} and DTIN-Net~\cite{DTIN-Net} align multiple domains through normalization techniques. Similarly, SVIL~\cite{SVIL} uses normalization to eliminate the effect of style factors across different domains. Additionally, Hu et al.~\cite{hu2024exert} introduce a large-scale DG Re-ID benchmark based on diverse feature space learning through normalization. Other methods like M$^3$L~\cite{M3L} and SuA-SpML~\cite{SuA-SpML} leverage meta-learning to acquire domain-invariant representations. MDA~\cite{MDA} is also based on meta-learning but employs test-time training simultaneously. The QAConv series~\cite{QAConv, QAConv-GS, QAConv-MS} and TransMatcher~\cite{TransMatcher} formulate DG Re-ID as a matching problem. In addition, approaches such as PAT~\cite{PAT}, GMN~\cite{GMN}, and ISTDG~\cite{ISTDG} focus on enhancing local features and improving model architectures. Methods like ISR~\cite{ISR}, DMRL~\cite{DMRL}, and ReMix~\cite{ReMiX} aim to boost generalization with more data. Different from these approaches, our work refines the inference stage of DG Re-ID task by re-ranking improvement rather than training a Re-ID encoder.

\subsection{MLLM-based Person Re-ID}
Recent years have witnessed the remarkable success of MLLMs in various vision-language tasks, owing to their powerful zero-shot generalization and semantic reasoning capabilities. Several pioneering studies have attempted to integrate MLLMs into Re-ID pipeline. For instance, LVLM-ReID~\cite{LVLM-ReID} and MLLMReID\cite{MLLMReID} explore the potential of using MLLMs as powerful feature extractors, leveraging their pre-trained knowledge to capture more robust person representations. Another line of research, represented by IRM~\cite{IRM} and ChatReID~\cite{ChatReID}, introduce an all-in-one paradigm where MLLMs are fine-tuned to perform different Re-ID tasks through natural language instructions within a single model. Unlike these approaches, our work acts as a powerful post-processing method by enhancing initial ranking results with an MLLM adapted to predict robust distances.

\subsection{MLLMs for Information Retrieval}
In Information Retrieval (IR), Large Language Models (LLMs) have been increasingly adopted to enhance textual content retrieval. Various paradigms have been explored, including pointwise~\cite{sachan2022improving, zhuang2023open}, listwise~\cite{Rankvicuna, ma2023zero, sun2023chatgpt}, pairwise~\cite{qin2024large}, and setwise~\cite{zhuang2024setwise} approaches, all demonstrating strong performance even under zero-shot inference settings. For multimodal retrieval, recent works such as MM-EMBED~\cite{MM-EMBED} and LamRA~\cite{LamRA} leverage MLLMs  to enable universal multimodal retrieval. MM-Embed~\cite{MM-EMBED}  employs continuous fine-tuning across multiple retrieval datasets and tasks, while LamRA~\cite{LamRA} adopts a two-stage training strategy comprising language-only pre-training and multimodal instruction tuning to improve retrieval performance. Differently, our work adapts an MLLM specifically for DG Re-ID through supervised fine-tuning, injecting domain-generalizable Re-ID knowledge and proposing a new distance metric to enhance re-ranking robustness.

\subsection{Re-Ranking for Person Re-ID}
Early work such as DaF~\cite{DaF} performs re-ranking with K-Nearest Neighbors (K-NN) following a divide-and-fuse strategy. ECN~\cite{ECN} improves upon K-NN by incorporating expanded samples to enhance matching accuracy. Zhong et al. subsequently introduce Jaccard distance encoding based on K-Reciprocal Nearest Neighbors (K-RNN)~\cite{CUHK03-NP}, which increases the true positive ratio within neighborhood sets. Zhang et al.. represent neighbors as graphs~\cite{zhang2020understanding} and employ Graph Neural Networks (GNNs) to effectively capture neighborhood information. Recent approaches~\cite{CAJ,CEIL} incorporate camera information into re-ranking. CAJ~\cite{CAJ} enhances Jaccard distance by considering both intra- and inter-camera neighbors, while CEIL~\cite{CEIL} combines camera information with GNN. Unlike these methods, our work leverages an MLLM to acquire an improved distance metric which is compatible with existing distance-based re-ranking methods for DG Re-ID.

\section{Recap of Re-Ranking for Person Re-ID}
Most re-ranking methods~\cite{CUHK03-NP,ECN,CAJ,DaF} in Re-ID are neighborhood-based and can be roughly summarized in following steps. First, an instance-level distance is computed as $D_{i,j} = \mathrm{Dist}(\bm{f}_i, \bm{f}_j)$, where $\mathrm{Dist}(\cdot, \cdot)$ is a distance metric, and $D_{i,j}$ denotes the distance between the image features $\bm{f}_i$ and $\bm{f}_j$. Second, a neighborhood set $\mathcal{N}_i = g(D_{i,:})$ is constructed for the $i$-th sample, where $D_{i,:}$ represents the distances between the $i$-th sample and all others, and $g(\cdot)$ is a neighborhood construction function. Third, a neighborhood-based distance $\tilde{D}_{i,j} = \Gamma(\mathcal{N}_i, \mathcal{N}_j)$ is computed, where $\Gamma(\cdot, \cdot)$ measures the similarity of the $i$-th and $j$-th samples according to their neighborhood sets. Finally, an improved ranking list is derived from $\tilde{D}_{i,j}$ to enhance Re-ID performance.

A critical step in these re-ranking methods is the accurate selection of positive neighbors. However, the neighborhood construction function fundamentally relies on the instance-level distances $D_{i,j}$, which are computed using image features extracted by the Re-ID encoder. Consequently, the effectiveness of re-ranking on target domains is inevitably constrained by the limited generalizability of the encoder itself.

\begin{figure*}[t]
	\centering
	\includegraphics[width=\textwidth]{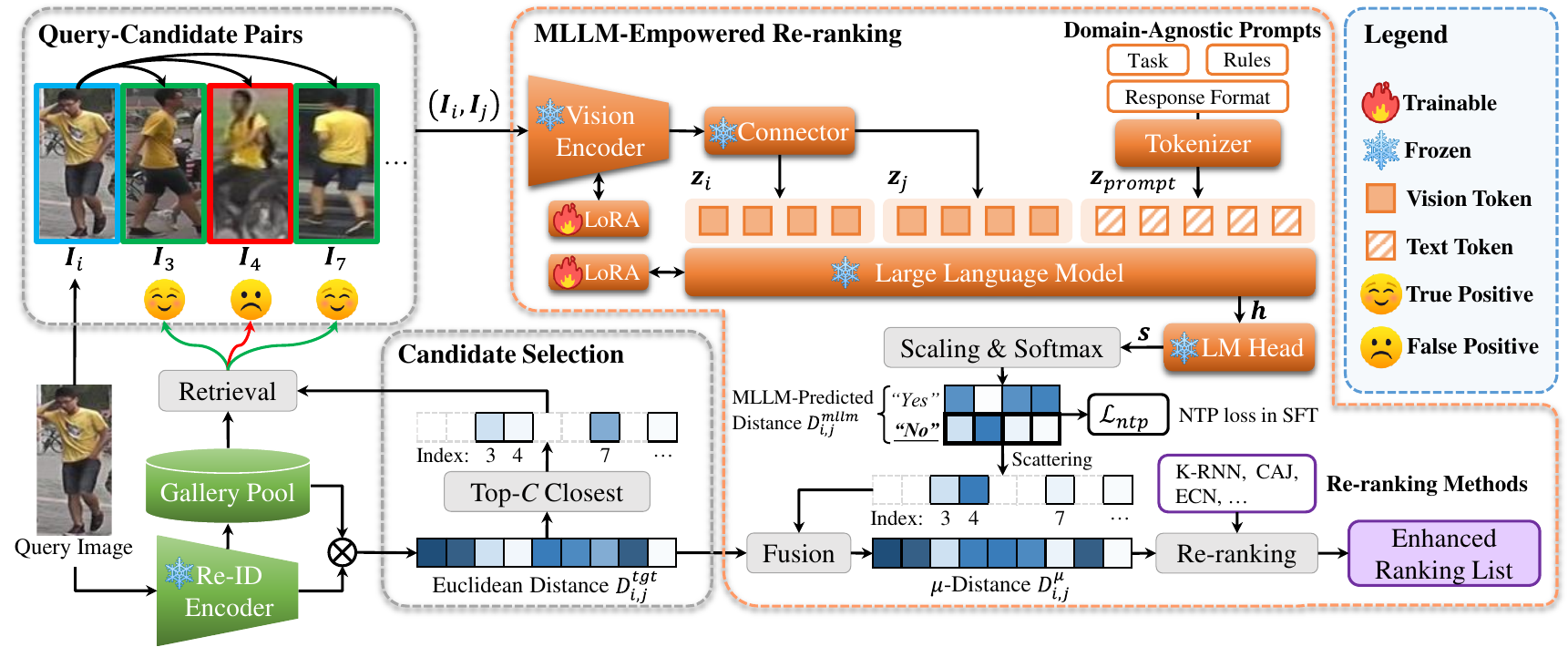}
	\caption{Pipeline of our approach. An Re-ID encoder is adopted for feature extraction and raw distance computation. An MLLM is adopted to rectify top-ranked retrievals based on the raw distance and produce a robust $\mu$-distance for re-ranking enhancement during inference.}
	\label{fig:pipeline}
\end{figure*}

\section{MLLM-Empowered Re-Ranking}

To address the aforementioned limitation of conventional re-ranking methods, we propose to leverage an MLLM to enhance the distance metric, making it more domain-agnostic and better suited for DG Re-ID re-ranking. The overall pipeline of our approach is presented in Figure~\ref{fig:pipeline}, which consists of a baseline Re-ID encoder for extracting image features and computing standard Euclidean distances, as well as a module providing an MLLM-predicted distances. These two distance metrics are then fused into a unified $\mu$-distance, which is further utilized for neighborhood construction and final re-ranking. In following sections, we will first briefly introduce the design of the baseline Re-ID encoder. After that, we will elaborate how to adapt the MLLM to enhance re-ranking.

\subsection{Baseline Re-ID Encoder}
\label{baseline}
Considering the widespread application of CLIP~\cite{CLIP} in Re-ID~\cite{CLIP-ReID,CLIP-DFGS,CLIP-FGDI}, we build a baseline Re-ID encoder following PCL-CLIP~\cite{PCL-CLIP} with prototypical contrastive loss. Given an input image, we extract its feature as $\bm{f}$ through the image encoder of CLIP. After that, the prototypical contrastive loss $\mathcal{L}_{pcl}$ is optimized to learn ID-discriminative features:
\begin{equation}
	\mathcal{L}_{pcl} = - \log \frac{\exp\left( \mathrm{sim}(\bm{f}, \mathcal{M}_y) / \tau_{reid} \right)}{\sum_{j=1}^{N_{id}} \exp\left( \mathrm{sim}(\bm{f}, \mathcal{M}_j) / \tau_{reid} \right)},
\end{equation}
where a temperature factor $\tau_{reid}$ is used to regulates the strength of contrasting. $y$ is the ID label of the feature. $\mathrm{sim}(\cdot, \cdot)$ denotes cosine similarity and $N_{id}$ denotes the total number of IDs. To maintain ID prototypes, a memory bank $\mathcal{M}$ is adopted to store the feature centroid per ID, which is updated in each training iteration with a momentum $\gamma$ using the hardest sample $\bm{f}^*$ labeled as $y$ in a batch~\cite{ClusterContrast}:
\begin{equation}
	\mathcal{M}_y \leftarrow \gamma \mathcal{M}_y + (1 - \gamma)\bm{f}^*.
\end{equation}
During the entire training process,  the classicial cross-entropy ID loss $\mathcal{L}_{id}$ and the prototypical contrastive loss $\mathcal{L}_{pcl}$ are optimized together to learn ID-discriminative feature extraction:
\begin{equation}
	\mathcal{L}_{reid} = \mathcal{L}_{id} + \mathcal{L}_{pcl}.
\end{equation}

\subsection{MLLM Adaptation on Re-ID Data}
\label{sec:muse}
MLLMs hold outstanding generalized knowledge across mulitple vision domains due to large-scale pre-training, but they still require further adaptation to perform better for specific tasks. Therefore, we design a binary matching task with a domain-agnostic prompt based on supervised fine-tuning paradigm to adapt the MLLM. Considering the effectiveness and efficiency of adaptation, the MLLM is fine-tuned with LoRA~\cite{LoRA} adapters and a query-candidate hard mining strategy.

\subsubsection{Domain-Agnostic Prompt}
The domain-agnostic prompt is adopted to inject ID-discriminative but domain-irrelated knowledge to the MLLM and enables query-candidate binary matching given a pair of input images. Specifically, we design a combined prompt composed of a task prompt $\mathcal{T}_{task}$, a rule prompt $\mathcal{T}_{rule}$, and a response format prompt $\mathcal{T}_{resp}$. These prompts are concatenated to form a Domain-Agnostic Prompt (DAP), denoted as $\mathcal{T}_{dap} = [\mathcal{T}_{task}, \mathcal{T}_{rule}, \mathcal{T}_{resp}]$, which is then fed into the MLLM.

The task prompt $\mathcal{T}_{task}$ defines the query-candidate matching task. It specifies the input format and assigns a role to the MLLM, clearly explaining its objective and expected behavior. The response format prompt $\mathcal{T}_{resp}$ is designed to constrain the output format of the MLLM. Specifically, the model is instructed to respond with either “Yes” or “No”, indicating whether the query and candidate images belong to the same identity.

The rule prompt $\mathcal{T}_{rule}$ is designed to define the criteria of ID discrimination and to mitigate the influence of domain-specific information. Specifically, we formulate a set of rules from multiple perspectives, including both ID-related and domain-related attributes. The former instructs the MLLM to attend to fine-grained visual cues relevant to human identity, while the latter reminds the model to disregard features that are domain-specific and irrelevant to ID matching.

Table~\ref{tab:prompt} shows all prompts used in our approach. The placeholders ``$\vartriangleleft$query$\vartriangleright$'', ``$\vartriangleleft$candidate$\vartriangleright$'', and ``$\vartriangleleft$ans$\vartriangleright$'' are replaced with query image tokens, candidate image tokens, and expected answer (``Yes'' or ``No'') during the adaptation, respectively. During inference, the ``$\vartriangleleft$ans$\vartriangleright$'' is removed and the MLLM generates a word as answer. Other task-irrelated system prompts are ignored for conciseness.

\begin{table*}[t]
	\centering
	\caption{Details of the prompts used in our approach.}
	\label{tab:prompt}
	\begin{tabularx}{\textwidth}{lX}
		\toprule
		Prompt type & \multicolumn{1}{c}{Content} \\
		\midrule
		Task prompt $\mathcal{T}_{task}$ & \makecell[l]{You are an expert in person recognition. I will give you a query\\ and a candidate image, captioned by ``Query:'' and ``Candidate:''\\ independently. Please tell me whether the people in the two\\ images have the same identity. Query:$\vartriangleleft$query$\vartriangleright$ Candidate:\\$\vartriangleleft$candidate$\vartriangleright$} \\
		\midrule
		Rule prompt $\mathcal{T}_{rule}$ &  \makecell[l]{You should distinguish the people identities based on following\\ criteria: \\1. appearance (e.g. age, gender, hair style, hair color, face, body\\ shape, etc.) \\ 2. clothing and wearing (e.g. clothing type, color, style, pattern,\\ etc.) \\ 3. carried objects (e.g. bags, hats, glasses, accessories, etc.) \\ Please keep in mind that you should never rely on following\\ attributes: \\ 1. background (e.g. location, time, etc.) \\ 2. human pose (e.g. standing, sitting, walking, etc.) \\ 3. occlusion (e.g. covered face, covered body, etc.) \\ 4. camera angle (e.g. front view, side view, back view, etc.)} \\
		\midrule
		Response format prompt $\mathcal{T}_{resp}$ & \makecell[l]{Respond with ``Yes'' if the query and the candidate images have\\ the same identity, otherwise respond with ``No''.$\vartriangleleft$ans$\vartriangleright$} \\
		\bottomrule
	\end{tabularx}
\end{table*}

\subsubsection{Supervised Fine-Tuning on Re-ID Dataset}

We employ supervised fine-tuning to adapt the MLLM, equipped with LoRA adapters~\cite{LoRA}, to Re-ID task. Specifically, a query image $\bm{I}_i$ and a candidate image $\bm{I}_j$ are sampled from the training set of a Re-ID dataset to form an image pair, which is then inserted into the task prompt described earlier. Each pair is labeled as either positive or negative depending on whether the two images share the same identity. Then, the image pair and the domain-agnostic prompt are encoded into latent embeddings $\bm{z}_i$, $\bm{z}_j$, and $\bm{z}_{prompt}$ via the MLLM’s vision encoder and text tokenizer. These embeddings are then concatenated into a sequence $\bm{z} = [\bm{z}_i, \bm{z}_j, \bm{z}_{prompt}]$, which is fed into the LLM decoder $\phi_{llm}(\cdot)$. The output embedding from the last decoder layer is obtained as $\bm{h} = \phi_{llm}(\bm{z})$. Finally, a language modeling head $\phi_{lm}(\cdot)$ is appended to predict the next-token logits as $\bm{s} = \phi_{lm}(\bm{h})$.

Assume that the sequence $\bm{z}$ consists of $L$ tokens, each corresponding to a word $x^l$ in the input prompt. The final word $x^L$ is either ``Yes'' or ``No'', depending on the label of the paired image input. The training objective is to maximize the log-likelihood of the final word conditioned on the preceding $L-1$ words, which is implemented using the Next-Token Prediction (NTP) loss $\mathcal{L}_{ntp}$:
\begin{equation}
	\begin{aligned}
		&P(x^{L} \ | \ x^{1:L-1}) = \mathrm{Softmax}(\bm{s}^{L-1}), \\
		&\mathcal{L}_{ntp} = - \log P(x^{L} \ | \ x^{1:L-1}),
	\end{aligned}
\end{equation}
where $\bm{s}^{L-1}$ is the logit corresponding to the $(L-1)$-th token, which is used to predict the final token. $P(x^{L} \ | \ x^{1:L-1})$ denotes the predicted probability distribution of the answer ``Yes'' or ``No'' at the $L$-th word, conditioned on all preceding tokens. 

\subsubsection{Query-Candidate Hard Mining}

To construct an input pair for a query image, a straightforward way is to randomly sample candidate images from the entire training set, excluding the query itself. However, most randomly sampled pairs are relatively easy to distinguish and offer limited learning value. To address this, we apply a Query-Candidate Hard Mining (QCHM) strategy that focuses training on hard query-candidate pairs, thereby enhancing the discriminative capability of the MLLM.

For all images in the training set, we extract their features using the baseline Re-ID encoder and compute the raw Euclidean distances of each pair. For the $i$-th query image, we select the most distant $C_{pos}$ images of the same identity as hard positive candidates, and the closest $C_{neg}$ images with different identities as hard negative candidates, based on the raw distance. We denote the positive and negative index sets of the $i$-th image as $\mathcal{P}_i$ and $\mathcal{Q}_i$, respectively. Using these hard samples, we construct a query-candidate hard mining dataset $\mathcal{D}_{QCHM} = \{(\bm{I}_{i}, \bm{I}_{j}) \ | \ j \in \mathcal{P}_i \cup \mathcal{Q}_i\}_{i=1}^N$, where $N$ is the number of samples. During training, image pairs are sampled from $\mathcal{D}_{QCHM}$ instead of the entire dataset, allowing the MLLM to focus on the most challenging examples.

\subsection{MLLM-Enhanced Distance Metric for Re-Ranking}
\label{sec:reranking_enhancement}
Neighborhood-based re-ranking is traditionally employed as a post-processing step during inference to enhance overall Re-ID accuracy. Given the testing set of target domain, the Euclidean distance $D^{tgt}_{i,j}$ is first computed based on image features extracted by the baseline Re-ID encoder. To improve the reliability of similarity estimation, we then refine the distance by fusing the original Euclidean distance with the MLLM-predicted distance. Finally, the refined distance is used to perform conventional neighborhood-based re-ranking.

Specifically, based on $D^{tgt}_{i,j}$, we select the top $C$ nearest retrievals as candidates for the $i$-th query image. Each query-candidate pair is then combined with the prompt $\mathcal{T}_{dap}$ and fed into the MLLM to produce a response of ``Yes'' or ``No''. To quantitatively measure the distance between the query and each candidate, we utilize the prediction logits of ``Yes'' or ``No'', inspired by pointwise likelihood generation~\cite{nogueira2020document}. That is, the MLLM-predicted distance is defined by
\begin{equation}
	D^{mllm}_{i,j} = \frac{\exp (s_{no}(\bm{I}_i, \bm{I}_j) / \tau)}{\exp (s_{yes}(\bm{I}_i, \bm{I}_j) / \tau) + \exp (s_{no}(\bm{I}_i, \bm{I}_j) / \tau)},
\end{equation}
where the subscript $j$ denotes the index of each candidate. $s_{yes}(\bm{I}_i, \bm{I}_j)$ and $s_{no}(\bm{I}_i, \bm{I}_j)$ are the output logits of ``Yes'' and ``No''. A temperature factor $\tau$ is used to scale the confidence of the prediction.

Although the MLLM-predicted distance $D^{mllm}_{i,j}$ is more generalizable across unseen domains, it is not reliable to use it alone, as MLLMs may inevitably suffer from hallucinations~\cite{HALoGEN,huang2023survey}. Therefore, we propose a new metric, termed the $\mu$-distance, which fuses the raw Euclidean distance with the MLLM-based distance. This fusion integrates the complementary strengths of both metrics, resulting in a distance metric that is more accurate and robuster to domain shifts. Given the $i$-th query image and the $j$-th image in the gallery set, the $\mu$-distance $D^{\mu}_{i,j}$ is defined as:
\begin{equation}
	D^{\mu}_{i,j} =
	\begin{cases}
		(1 - \alpha) D^{tgt}_{i, j} + \alpha D^{mllm}_{i, j} & \text{if\ } j \in \mathcal{C}_i \\
		D^{tgt}_{i,j} & \text{otherwise}
	\end{cases},
\end{equation}
where $j \in \{1,\dots, G\}$ and $G$ denote the size of the gallery set. $\mathcal{C}_i$ represents the index set of top-ranked candidates for the $i$-th query, determined based on the Euclidean distance. If a gallery image indexed by $j$ appears among these top-ranked candidates, its Euclidean distance and MLLM-predicted distance are linearly combined, with a weighting factor $\alpha$ controlling fusion strength. This fusion is restricted to the top-ranked candidates to reduce the computational overhead of using the MLLM to evaluate additional query-candidate pairs that are unlikely to contribute meaningfully to the subsequent re-ranking process.

Finally, conventional re-ranking methods such as K-RNN~\cite{CUHK03-NP}, ECN~\cite{ECN} and CAJ~\cite{CAJ} can be applied based on neighborhoods constructed according to the $\mu$-distance $D^{\mu}_{i,j}$. 

\section{Experiments}
\label{sec:exp}

\subsection{Evaluation Protocols}

We conduct experiments on Re-ID datasets Market1501~\cite{Market1501}, MSMT17~\cite{MSMT17}, CUHK03-NP~\cite{CUHK03-NP} and CUHK-SYSU~\cite{CUHK-SYSU}, abbreviated as MA, MS, C3, and CS, respectively. The DG Re-ID is evaluated under both single-source and multi-source protocols. The former trains the Re-ID encoder on the training set of one dataset and evaluated on the testing set of another dataset. The latter adopts a leave-one-out strategy to test on one dataset while train on the mixture of the training sets of remaining datasets. Following previous approaches~\cite{CLIP-DFGS,CLIP-FGDI}, CS is only employed for multi-source training and not used for testing. The Mean Average Precision (mAP) and Cumulative Matching Characteristic (CMC) at Rank-1 are reported.

\subsection{Implementation Details}

The baseline Re-ID encoder follows the training details from PCL-CLIP~\cite{PCL-CLIP} but takes a batch size of 32 under a learning rate of $5\times 10^{-6}$ without random cropping and erasing~\cite{random_erasing} augmentations. For the MLLM, we select Qwen2-VL-2B~\cite{Qwen2VL} architecture. The input resolution of image is set to 280$\times$140. We fine-tune the linear layers in the vision encoder and the LLM decoder of the MLLM with LoRA~\cite{LoRA} adapters of rank 16. The number of hard samples $C_{pos}$ and $C_{neg}$ are both set to 5 in QCHM. In inference, the number of candidates $C$ is set to 40 with $\tau=5$. The $\mu$-distance is obtained by fusion rate $\alpha=0.2$. We adopt the AdamW~\cite{AdamW} optimizers in both baseline training and MLLM fine-tuning. The MLLM is optimized with a learning rate of $5\times 10^{-5}$ for 20 epochs, regulated by a cosine annealing scheduler~\cite{cosine_annealing} with a batch size of 2. We conduct the experiments on 4 NVIDIA RTX A6000 GPUs with BF16 precision.

\begin{table*}[t]
	\centering
	\caption{Comparison with state-of-the-art re-ranking methods on single-source generalization. Best results are emphasized in bold and second-best results are highlighted with underline.}
	\label{tab:sft_ma_ssdg}
	\resizebox{0.9\textwidth}{!}{
		\begin{tabular}{lccccccccc}
			\toprule
			\multirow{2}{*}{Model} & \multirow{2}{*}{Re-ranking} & \multicolumn{2}{c}{MA$\to$MS} & \multicolumn{2}{c}{MA$\to$C3} & \multicolumn{2}{c}{MS$\to$C3} & \multicolumn{2}{c}{C3$\to$MS} \\ \cmidrule{3-10} 
			&  & mAP & Rank-1 & mAP & Rank-1 & mAP & Rank-1 & mAP & Rank-1 \\ \midrule
			STL~\cite{STL}~$_{\textit{ICME'24}}$ & \XSolidBrush & 19.8 & 48.9 & 26.9 & 27.7 & 24.5 & 25.6 & - & - \\
			STL+PAT~\cite{STL}~$_{\textit{ICME'24}}$ & \XSolidBrush & 19.8 & 48.9 & 26.9 & 27.7 & 24.5 & 25.6 & - & - \\
			LDU~\cite{LDU}~$_{\textit{TIM'24}}$ & \XSolidBrush & 13.5 & 35.7 & 18.2 & 18.5 & 21.3 & 21.3 & 12.6 & 36.9 \\
			QAConv-MS~\cite{QAConv-MS}~$_{\textit{TCSVT'24}}$ & \XSolidBrush & 19.9 & 49.7 & 24.9 & 26.6 & 28.5 & 31.0 & - & - \\
			DMRL~\cite{DMRL}~$_{\textit{Mach. Learn.'24}}$ & \XSolidBrush & 21.5 & 50.6 & 22.6 & 23.4 & 24.7 & 26.1 & - & - \\
			DCAC~\cite{DCAC}~$_{\textit{Sensors'25}}$ & \XSolidBrush & 23.4 & 52.1 & 32.5 & 33.2 & 34.1 & 34.4 & 17.8 & 47.3 \\
			CLIP-FGDI~\cite{CLIP-FGDI}~$_{\textit{TIFS'25}}$ & \XSolidBrush & 20.2 & 47.0 & 33.2 & 34.6 & 30.1 & 32.4 & 17.7 & 45.0 \\
			\midrule
			Baseline~$_{\textit{Ours}}$ & \XSolidBrush & 24.7 & 53.4 & 36.9 & 37.1 & 35.0 & 36.5 & 23.7 & 54.4 \\
			Baseline + K-RNN~\cite{CUHK03-NP}~$_{\textit{CVPR'17}}$ & \Checkmark & 34.5 & 58.5 & 51.8 & 45.9 & 49.5 & 44.9 & 35.9 & 61.7 \\
			Baseline + ECN~\cite{ECN}~$_{\textit{CVPR'18}}$ & \Checkmark & 41.4 & 60.6 & 50.3 & 46.1 & 47.4 & 43.3 & 44.6 & 64.2 \\
			Baseline + CAJ~\cite{CAJ}~$_{\textit{CVPR'24}}$ & \Checkmark & 40.0 & 60.9 & 53.8 & 49.9 & 52.1 & 47.7 & 44.5 & 66.1 \\ \midrule
			Baseline + MUSE~$_{\textit{Ours}}$ & \XSolidBrush & 27.4 & 59.3 & 42.4 & 45.3 & 42.5 & 47.7 & 27.3 & 61.6 \\
			Baseline + MUSE + K-RNN~$_{\textit{Ours}}$ & \Checkmark & 37.5 & 62.7 & \underline{56.8} & \underline{51.7} & \underline{56.1} & \underline{51.4} & 39.6 & 66.9 \\
			Baseline + MUSE + ECN~$_{\textit{Ours}}$ & \Checkmark & \textbf{44.7} & \textbf{64.4} & 55.1 & 50.8 & 54.4 & 50.3 & \textbf{48.5} & \underline{68.7} \\
			Baseline + MUSE + CAJ~$_{\textit{Ours}}$ & \Checkmark & \underline{41.8} & \underline{63.9} & \textbf{57.3} & \textbf{53.4} & \textbf{57.2} & \textbf{53.9} & \underline{46.4 }& \textbf{68.8} \\
			\bottomrule
	\end{tabular}}
\end{table*}

\begin{table*}[t]
	\centering
	\caption{Comparison with state-of-the-art re-ranking methods on multi-source generalization. Best results are emphasized in bold and second-best results are highlighted with underline.}
	\label{tab:sft_ma_msdg}
	\resizebox{0.75\textwidth}{!}{
		\begin{tabular}{lccccc}
			\toprule
			\multirow{2}{*}{Model} & \multirow{2}{*}{Re-ranking} & \multicolumn{2}{c}{MA+MS+CS$\to$C3} & \multicolumn{2}{c}{MA+C3+CS$\to$MS} \\ \cmidrule{3-6} 
			& & mAP & Rank-1 & mAP & Rank-1 \\ \midrule
			ReNorm~\cite{ReNorm}~$_{\textit{ECCV'24}}$ & \XSolidBrush & 43.6 & 44.7 & 25.6 & 55.6 \\
			DFGS~\cite{CLIP-DFGS}~$_{\textit{TOMM'24}}$ & \XSolidBrush & 50.4 & 51.1 & 31.5 & 59.7 \\
			CLIP-FGDI~\cite{CLIP-FGDI}~$_{\textit{TIFS'25}}$ & \XSolidBrush & 44.4 & 44.6 & 31.1 & 59.4 \\ \midrule
			Baseline~$_{\textit{Ours}}$ & \XSolidBrush & 39.6 & 42.8 & 21.9 & 48.5 \\
			Baseline + K-RNN~\cite{CUHK03-NP}~$_{\textit{CVPR'17}}$ & \Checkmark & 55.5 & 51.9 & 32.6 & 55.8 \\
			Baseline + ECN~\cite{ECN}~$_{\textit{CVPR'18}}$ & \Checkmark & 53.6 & 50.0 & 41.6 & 58.8 \\
			Baseline + CAJ~\cite{CAJ}~$_{\textit{CVPR'24}}$ & \Checkmark & 56.1 & 52.6 & 43.9 & 63.4 \\ \midrule
			Baseline + MUSE~$_\textit{Ours}$ & \XSolidBrush & 44.1 & 48.7 & 25.6 & 57.7 \\
			Baseline + MUSE + K-RNN~$_\textit{Ours}$ & \Checkmark & \textbf{59.6} & \underline{54.9} & 35.0 & 59.4 \\
			Baseline + MUSE + ECN~$_\textit{Ours}$ & \Checkmark & 58.5 & 54.6 & \underline{46.5} & \underline{64.0} \\
			Baseline + MUSE + CAJ~$_\textit{Ours}$ & \Checkmark & \underline{59.4} & \textbf{55.7} & \textbf{46.7} & \textbf{66.6} \\ \bottomrule
	\end{tabular}}
\end{table*}

\begin{table*}[tp]
	\centering
	\caption{Re-ranking performances on source domains. Best results are emphasized in bold and second-best results are highlighted with underline.}
	\label{tab:intra_domain_perf}
	\resizebox{0.85\textwidth}{!}{
		\begin{tabular}{lccccccc}
			\toprule
			\multirow{2}{*}{Model} & \multirow{2}{*}{Re-ranking} & \multicolumn{2}{c}{Market1501} & \multicolumn{2}{c}{CUHK03-NP} & \multicolumn{2}{c}{MSMT17} \\ \cmidrule{3-8} 
			& & mAP & Rank-1 & mAP & Rank-1 & mAP & Rank-1 \\ \midrule
			Baseline~$_\textit{Ours}$ & \XSolidBrush & 87.5 & 94.7 & 69.0 & 72.8 & 69.7 & 87.7 \\
			Baseline + K-RNN~\cite{CUHK03-NP}~$_\textit{CVPR'17}$ & \Checkmark & 91.7 & 94.8 & 85.5 & 82.9 & 76.8 & 87.8 \\
			Baseline + ECN~\cite{ECN}~$_\textit{CVPR'18}$ & \Checkmark & 93.5 & \underline{95.6} & 84.4 & 83.1 & 83.7 & 89.9 \\
			Baseline + CAJ~\cite{CAJ}~$_\textit{CVPR'24}$ & \Checkmark & 93.6 & 95.3 & 84.8 & 82.0 & 83.4 & 90.0 \\ \midrule
			Baseline + MUSE~$_\textit{Ours}$ & \XSolidBrush & 89.4 & 95.1 & 82.3 & 86.2 & 70.6 & 87.8 \\
			Baseline + MUSE + K-RNN~$_\textit{Ours}$ & \Checkmark & 92.7 & 95.5 & \textbf{90.5} & \textbf{88.4} & 77.5 & 87.7 \\
			Baseline + MUSE + ECN~$_\textit{Ours}$ & \Checkmark & \underline{94.1} & \underline{95.6} & 89.0 & \textbf{88.4} & \textbf{84.6} & \textbf{90.2} \\
			Baseline + MUSE + CAJ~$_\textit{Ours}$ & \Checkmark & \textbf{94.4} & \textbf{95.8} & \underline{89.2} & \underline{87.9} & \underline{84.1} & \underline{90.1} \\ \bottomrule
	\end{tabular}}
\end{table*}

\subsection{Comparison with State-of-the-Arts}
\label{sec:exp_sota}

\subsubsection{Performance on DG Re-ID}
\label{sec:perf_dg}
We first evaluate the performance on single-source generalization. Note that testing on the same domain used in MLLM adaptation may cause potential domain-leakage problem, which is unfair to evaluate the model's actual capability since it may contact target domain's knowledge in advance. Thus, the MLLM is merely fine-tuned on MA's training set and utilized in all evaluation protocols except those taking MA as target domain. If you are interested in the results collected when adaptation and testing are carried on the same domain, please refer to Appendix~\ref{appendix:perf_tgt_MA} for more details.

As shown in Table~\ref{tab:sft_ma_ssdg}, the baseline has already achieved good results before re-ranking. Multiple re-ranking methods, including K-RNN~\cite{CUHK03-NP}, ECN~\cite{ECN} and CAJ~\cite{CAJ}, contribute further improvements on the baseline. When our approach (denoted as ``MUSE'') is employed, consistent improvements can be observed over all generalization tests even if we directly use the $\mu$-distance to compute Re-ID results (denoted as ``Baseline + MUSE'') without neighborhood-based re-ranking methods. This indicates that the raw distance is indeed enhanced. Moreover, when the re-ranking methods are adopted based on our $\mu$-distance, the performance is boosted further, showing the effectiveness of our approach.

Results from Table~\ref{tab:sft_ma_msdg} on multi-source generalization provide with more evidences. Without re-ranking, the baseline presents obvious performance gaps over all generalization tests. Existing re-ranking methods are able to apparently mitigate the gaps but the improvements are limited. By adopting the MLLM-based distance enhancement, the performance of the baseline surpasses previous results re-ranking with raw distance by considerable margins.

\subsubsection{Performance on Intra-Domain Re-ID}
\label{sec:perf_intra}
Our approach is also applicable to intra-domain Re-ID. Different from DG Re-ID, it is less likely to be affected by domain gap since the training and testing sets share the same distribution. Thus, the improvement brought by the generalized knowledge introduced by MLLM is expected to be less pronounced than that in DG Re-ID. To verify this point, we report source domain performance in Table~\ref{tab:intra_domain_perf}. Unlike the DG scenario, the re-ranking performance on source domain is already relatively high, making it more susceptible to the hallucination of MLLM. For this reason, we reduce the fusion weight $\alpha$ to a more moderate value of 0.1. Experimental results demonstrate that the proposed approach still consistently improves the re-ranking performance on source domains. Nevertheless, since there is no significant domain gap between training and testing phases, the improvements yielded by traditional re-ranking have already saturated. As a result, the enhancement effect of MLLM is less prominent compared with that in DG Re-ID, which is consistent with our expectation. A notable exception is the result on the CUHK03-NP dataset. Owing to its inherently low baseline performance, the proposed approach can still bring substantial improvements. Generally speaking, results in Section~\ref{sec:perf_dg} and \ref{sec:perf_intra} reveal that our approach indeed refines the generalization capabilities of re-ranking methods, and preserves their effectiveness on source domains at the same time.

\subsection{Ablation Studies}
\subsubsection{Effectiveness of DAP}
\begin{table}[tb]
	\centering
	\caption{Ablations on domain-agnostic prompt composition. Best results are emphasized in bold.}
	\label{tab:dap}
	\resizebox{\linewidth}{!}{
		\begin{tabular}{lcccc}
			\toprule
			\multirow{2}{*}{Model} & \multicolumn{2}{c}{MA$\to$C3} & \multicolumn{2}{c}{MA$\to$MS} \\ \cmidrule{2-5} 
			& mAP & Rank-1 & mAP & Rank-1 \\ \midrule
			Baselinse + K-RNN~\cite{CUHK03-NP} & 51.8 & 45.9 & 34.5 & 58.5 \\
			+ MUSE (Void prompt) & 56.3 & 50.9 & 36.6 & 62.2 \\
			+ MUSE ($\mathcal{T}_{task}$ + $\mathcal{T}_{resp}$) & 56.2 & 50.7 & 36.4 & 61.8 \\
			+ MUSE ($\mathcal{T}_{task}$ + $\mathcal{T}_{resp}$ + $\mathcal{T}_{rule})$ & \textbf{56.8} & \textbf{51.7} & \textbf{37.5} & \textbf{62.7} \\ \bottomrule
	\end{tabular}}
\end{table}

\begin{table}[tb]
	\centering
	\caption{Effectiveness of adaptation. Best results are emphasized in bold.}
	\label{tab:sft}
	\resizebox{\linewidth}{!}{
		\begin{tabular}{lcccc}
			\toprule
			\multirow{2}{*}{Re-ranking method} & \multicolumn{2}{c}{MA$\to$C3} & \multicolumn{2}{c}{MA$\to$MS} \\ \cmidrule{2-5} 
			& mAP & Rank-1 & mAP & Rank-1\\
			\midrule
			Baseline + K-RNN~\cite{CUHK03-NP} & 51.8 & 45.9 & 34.5 & 58.5 \\
			+ MUSE (Qwen2-VL-2B~\cite{Qwen2VL}, zero-shot) & 51.3 & 45.7 & 32.3 & 53.2 \\
			+ MUSE (Qwen2.5-VL-7B~\cite{Qwen2.5VL}, zero-shot) & 51.9 & 45.9 & 34.8 & 58.4 \\
			+ MUSE (Qwen2-VL-2B~\cite{Qwen2VL}, SFT) & \textbf{56.8} & \textbf{51.7} & 37.5 & 62.7 \\
			+ MUSE (Qwen2.5-VL-7B~\cite{Qwen2.5VL}, SFT) & 56.3 & 50.6 & \textbf{37.6} & \textbf{62.8} \\
			\bottomrule
	\end{tabular}}
\end{table}

\begin{table}[tb]
	\centering
	\caption{Effectiveness of query-candidate hard mining. Best results are emphasized in bold.}
	\label{tab:qchm}
	\resizebox{\linewidth}{!}{
		\begin{tabular}{lcccc}
			\toprule
			\multirow{2}{*}{Model} & \multicolumn{2}{c}{MA$\to$C3} & \multicolumn{2}{c}{MA$\to$MS} \\
			\cmidrule{2-5} 
			& mAP & Rank-1 & mAP & Rank-1 \\
			\midrule
			Baseline + K-RNN~\cite{CUHK03-NP} & 51.8 & 45.9 & 34.5 & 58.5 \\
			+ MUSE (Random sampling) & 51.2 & 46.1 & 34.8 & 59.2 \\
			+ MUSE (QCHM) & \textbf{56.8} & \textbf{51.7} & \textbf{37.5} & \textbf{62.7} \\
			\bottomrule
	\end{tabular}}
\end{table}

In Table~\ref{tab:dap}, we study the effectiveness of the composition of the Domain-Agnostic Prompt (DAP). When no prompts are used, denoted as ``+ MUSE (Void prompt)'', the adaptation fine-tuning is able to bring a basic performance improvement by only focusing on input images. Interestingly, we find the performance is slightly impaired when we only adopt the task and the response format prompts. In this case, the MLLM may exploit the knowledge that is not the most important for DG Re-ID to complete the task as instructed. When the rule prompt is added to explicitly specify the criteria of ID discrimination, the correct knowledge is injected to obtain the best performance.

\begin{table*}[tb]
	\centering
	\caption{Re-ranking performances of source-aware adaptation on single-source generalization. The MLLM is fine-tuned on the source domain for Re-ID encoder training. Best results are emphasized in bold and second-best results are highlighted with underline.}
	\label{tab:source_aware_sft_ssdg}
	\resizebox{\textwidth}{!}{
		\begin{tabular}{lccccccccc}
			\toprule
			\multirow{2}{*}{Model} & \multirow{2}{*}{Re-ranking} & \multicolumn{2}{c}{MS$\to$MA} & \multicolumn{2}{c}{MS$\to$C3} & \multicolumn{2}{c}{C3$\to$MA} & \multicolumn{2}{c}{C3$\to$MS} \\ \cmidrule{3-10} 
			&  & mAP & Rank-1 & mAP & Rank-1 & mAP & Rank-1 & mAP & Rank-1 \\ \midrule
			Baseline~$_{\textit{Ours}}$ & \XSolidBrush & 50.9 & 76.8 & 35.0 & 36.5 & 51.8 & 75.3 & 23.7 & 54.4 \\
			Baseline + K-RNN~\cite{CUHK03-NP}~$_{\textit{CVPR'17}}$ & \Checkmark & 63.0 & 77.1 & 49.5 & 44.9 & 66.7 & 78.7 & 35.9 & 61.7 \\
			Baseline + ECN~\cite{ECN}~$_{\textit{CVPR'18}}$ & \Checkmark & 68.4 & 78.9 & 47.4 & 43.3 & 71.1 & 80.5 & 44.6 & 64.2 \\
			Baseline + CAJ~\cite{CAJ}~$_{\textit{CVPR'24}}$ & \Checkmark & 68.1 & 79.3 & 52.1 & \underline{47.7} & 74.4 & 82.1 & 44.5 & 66.1 \\ \midrule
			Baseline + MUSE~$_{\textit{Ours}}$ & \XSolidBrush & 64.3 & 83.0 & 38.5 & 41.4 & 58.8 & 81.9 & 27.6 & 64.4 \\
			Baseline + MUSE + K-RNN~$_{\textit{Ours}}$ & \Checkmark & 73.8 & 82.1 & \underline{52.8} & 47.1 & 74.3 & 83.2 & 40.9 & 68.1 \\
			Baseline + MUSE + ECN~$_{\textit{Ours}}$ & \Checkmark & \textbf{78.4} & \textbf{85.2} & 51.7 & 47.3 & \underline{77.9} & \underline{85.3} & \textbf{49.8} & \underline{69.8} \\
			Baseline + MUSE + CAJ~$_{\textit{Ours}}$ & \Checkmark & \underline{78.2} & \underline{84.7} & \textbf{54.1} & \textbf{50.6} & \textbf{79.6} & \textbf{86.5} & \underline{48.6} & \textbf{70.8} \\
			\bottomrule
	\end{tabular}}
\end{table*}

\begin{table*}[tb]
	\centering
	\caption{Re-ranking performances of source-aware adaptation on multi-source generalization. The MLLM is fine-tuned on the source domains for Re-ID encoder training. Best results are emphasized in bold and second-best results are highlighted with underline.}
	\label{tab:source_aware_sft_msdg}
	\resizebox{0.9\textwidth}{!}{
		\begin{tabular}{lccccccc}
			\toprule
			\multirow{2}{*}{Model} & \multirow{2}{*}{Re-ranking} & \multicolumn{2}{c}{MA+MS+CS$\to$C3} & \multicolumn{2}{c}{MS+C3+CS$\to$MA} & \multicolumn{2}{c}{MA+C3+CS$\to$MS} \\ \cmidrule{3-8} 
			& & mAP & Rank-1 & mAP & Rank-1 & mAP & Rank-1 \\ \midrule
			Baseline~$_{\textit{Ours}}$ & \XSolidBrush & 39.6 & 42.8 & 62.3 & 81.4 & 21.9 & 48.5 \\
			Baseline + K-RNN~\cite{CUHK03-NP}~$_{\textit{CVPR'17}}$ & \Checkmark & 55.5 & 51.9 & 74.3 & 82.8 & 32.6 & 55.8 \\
			Baseline + ECN~\cite{ECN}~$_{\textit{CVPR'18}}$ & \Checkmark & 53.6 & 50.0 & 78.9 & 84.7 & 41.6 & 58.8 \\
			Baseline + CAJ~\cite{CAJ}~$_{\textit{CVPR'24}}$ & \Checkmark & 56.1 & 52.6 & \underline{81.9} & 86.7 & 43.9 & 63.4 \\ \midrule
			Baseline + MUSE~$_{\textit{Ours}}$ & \XSolidBrush & 43.8 & 47.7 & 57.7 & 83.2 & 25.9 & 61.8 \\
			Baseline + MUSE + K-RNN~$_{\textit{Ours}}$ & \Checkmark & \textbf{60.0} & \underline{55.4} & 76.4 & 84.5 & 38.7 & 64.3 \\
			Baseline + MUSE + ECN~$_{\textit{Ours}}$ & \Checkmark & 58.9 & 55.2 & 81.8 & \underline{86.9} & \underline{48.2} & \underline{66.3} \\
			Baseline + MUSE + CAJ~$_{\textit{Ours}}$ & \Checkmark & \underline{59.8} & \textbf{56.0} & \textbf{83.8} & \textbf{87.7} & \textbf{48.8} & \textbf{69.2} \\ \bottomrule
	\end{tabular}}
\end{table*}

\subsubsection{Effectiveness of Adaptation}

To validate the necessity of MLLM adaptation, we first evaluate the MLLM-empowered re-ranking through zero-shot inference. As presented in Table~\ref{tab:sft}, we conduct ablations on models with different sizes, including Qwen2-VL-2B~\cite{Qwen2VL} and larger Qwen2.5-VL-7B~\cite{Qwen2.5VL} using more parameters and better architecture. Unfortunately, the zero-shot models present no apparent enhancements but only tiny flunctuations due to poor adaptation to DG Re-ID even if the domain-agnostic prompt is used to inject task-specific knowledge during inference. While the fine-tuned models achieve prominent enhancements, greatly indicating the importance of MLLM adaptation in this task. From the results of the 7B model, we find that more parameters do not further contribute significant refinement. Thus, we choose the smaller 2B model in this task.

\begin{table*}[tbp]
	\centering
	\caption{Re-ranking performances of source-agnostic adaptation on single-source generalization. The MLLM is fine-tuned on MS and tested on MA and C3. Best results are emphasized in bold and second-best results are highlighted with underline.}
	\label{tab:source_agnostic_sft_ms}
	\resizebox{0.7\textwidth}{!}{
		\begin{tabular}{lccccc}
			\toprule
			\multirow{2}{*}{Model} & \multirow{2}{*}{Re-ranking} & \multicolumn{2}{c}{MA$\to$C3} & \multicolumn{2}{c}{C3$\to$MA} \\ \cmidrule{3-6} 
			&  & mAP & Rank-1 & mAP & Rank-1 \\ \midrule
			Baseline~$_{\textit{Ours}}$ & \XSolidBrush & 36.9 & 37.1 & 51.8 & 75.3 \\
			Baseline + K-RNN~\cite{CUHK03-NP}~$_{\textit{CVPR'17}}$ & \Checkmark & 51.8 & 45.9 & 66.7 & 78.7 \\
			Baseline + ECN~\cite{ECN}~$_{\textit{CVPR'18}}$ & \Checkmark & 50.3 & 46.1 & 71.1 & 80.5 \\
			Baseline + CAJ~\cite{CAJ}~$_{\textit{CVPR'24}}$ & \Checkmark & 53.8 & \underline{49.9} & 74.4 & 82.1 \\ \midrule
			Baseline + MUSE~$_{\textit{Ours}}$ & \XSolidBrush & 39.1 & 41.3 & 56.0 & 78.2 \\
			Baseline + MUSE + K-RNN~$_{\textit{Ours}}$ & \Checkmark & \underline{53.9} & 49.4 & 69.2 & 80.6 \\
			Baseline + MUSE + ECN~$_{\textit{Ours}}$ & \Checkmark & 52.9 & 49.1 & \underline{74.6} & \underline{82.5} \\
			Baseline + MUSE + CAJ~$_{\textit{Ours}}$ & \Checkmark & \textbf{54.8} & \textbf{51.1} & \textbf{75.4} & \textbf{83.6} \\ \bottomrule
	\end{tabular}}
\end{table*}

\begin{table*}[tbp]
	\centering
	\caption{Re-ranking performances of source-agnostic adaptation on single-source generalization. The MLLM is fine-tuned on C3 and tested on MA and MS. Best results are emphasized in bold and second-best results are highlighted with underline.}
	\label{tab:source_agnostic_sft_c3}
	\resizebox{0.7\textwidth}{!}{
		\begin{tabular}{lccccc}
			\toprule
			\multirow{2}{*}{Model} & \multirow{2}{*}{Re-ranking} & \multicolumn{2}{c}{MA$\to$MS} & \multicolumn{2}{c}{MS$\to$MA} \\ \cmidrule{3-6} 
			&  & mAP & Rank-1 & mAP & Rank-1 \\ \midrule
			Baseline~$_{\textit{Ours}}$ & \XSolidBrush & 24.7 & 53.4 & 50.9 & 76.8 \\
			Baseline + K-RNN~\cite{CUHK03-NP}~$_{\textit{CVPR'17}}$ & \Checkmark & 34.5 & 58.5 & 63.0 & 77.1 \\
			Baseline + ECN~\cite{ECN}~$_{\textit{CVPR'18}}$ & \Checkmark & 41.4 & 60.6 & 68.4 & 78.9 \\
			Baseline + CAJ~\cite{CAJ}~$_{\textit{CVPR'24}}$ & \Checkmark & 40.0 & 60.9 & 68.1 & 79.3 \\ \midrule
			Baseline + MUSE~$_{\textit{Ours}}$ & \XSolidBrush & 27.7 & 61.6 & 57.6 & 83.0 \\
			Baseline + MUSE + K-RNN~$_{\textit{Ours}}$ & \Checkmark & 38.3 & 64.1 & 70.6 & 82.5 \\
			Baseline + MUSE + ECN~$_{\textit{Ours}}$ & \Checkmark & \textbf{45.9} & \underline{65.9} & \textbf{75.4} & \underline{84.1} \\
			Baseline + MUSE + CAJ~$_{\textit{Ours}}$ & \Checkmark & \underline{43.8} & \textbf{66.0} & \underline{75.1} & \textbf{84.6} \\ \bottomrule
	\end{tabular}}
\end{table*}

\subsubsection{Effectiveness of QCHM}

In Table~\ref{tab:qchm}, we study the effectiveness of the proposed Query-Candidate Hard Mining (QCHM) in Supervised Fine-Tuning (SFT). In ``+ MUSE (Random sampling)'', we randomly select $C_{pos} + C_{neg}$ samples for each query image, leading to an SFT dataset mainly formed by easy samples. Obviously, the random sampling fails to offer significant performance improvements and even slightly impairs mAP on MA$\to$C3. However, the QCHM strategy exceeds the random selection strategy by an apparent margin, proving that focusing on hard samples is crucial to make our approach effective.

\subsubsection{Robustness to Fine-Tuning Data Selection}
\label{more_sft_exp}
We conduct more experiments to validate whether the MLLM adaptation is robust to the selection of fine-tuning data. Specifically, we adopt two settings: 1) \textbf{source-aware adaptation} and 2) \textbf{source-agnostic adaptation}. The former fine-tunes the MLLM on the same domain for DG Re-ID training. But the latter fine-tunes it on a different domain which is not used for either DG Re-ID training or testing. Table~\ref{tab:source_aware_sft_ssdg} and \ref{tab:source_aware_sft_msdg} present the results after source-aware adaptation in single-source and multi-source generalization protocols, respectively.

Moreover, Table~\ref{tab:source_agnostic_sft_ms} and \ref{tab:source_agnostic_sft_c3} demonstrate the results in single-source generalization protocols after source-agnostic adaptation, where the domains for adaptation are MS and C3, respectively. Since we use leave-one-out strategy in multi-source generalization, source-agnostic adaptation in such protocol is unavailable.

Consistently, we observe improved performances in all experiments regardless of employing source-aware or source-agnostic adaptation. There is one exception in Table~\ref{tab:source_aware_sft_msdg} when our approach is directly used on the baseline without neighbor-based re-ranking (denoted as ``Baseline + MUSE'') in MS+C3+CS$\to$MA generalization, where the mAP drops due to inappropriate choice of the fusion rate $\alpha$. When we choose a smaller $\alpha = 0.1$, we find its mAP and Rank-1 are refined to 63.3\% and 84.1\%, respectively.

Overall, these results indicate that our approach is not sensitive to the selection of fine-tuning data, demonstrating its outstanding robustness.

\subsubsection{Parameter Analysis}

\begin{figure*}[tb]
	\centering
	\includegraphics[width=\textwidth]{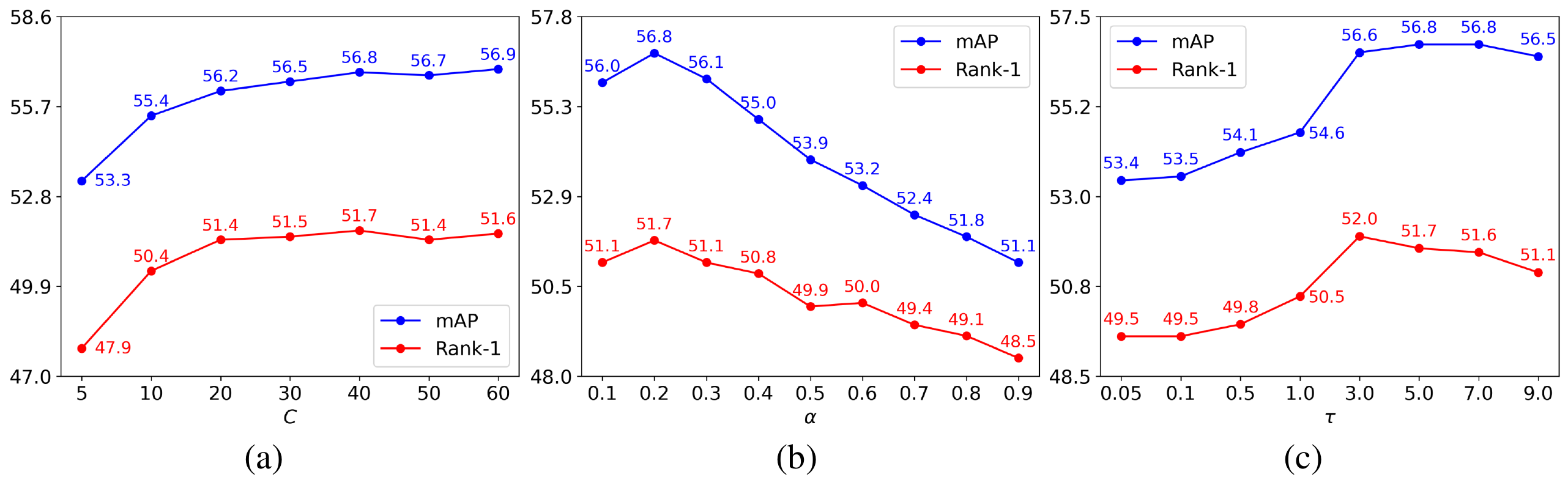}
	\caption{Parameter analysis on (a) the number of candidates $C$, (b) the fusion rate $\alpha$, and (c) the temperature $\tau$ during MLLM inference and distance enhancement.}
	\label{fig:vis_param_analysis}
\end{figure*}

On MA$\to$C3, we analyze three major hyper-parameters: $C$, $\alpha$, and $\tau$. $C$ determines how many top candidates from the raw distance are sent to the MLLM for rectification. As presented in Figure~\ref{fig:vis_param_analysis} (a), we choose $C=40$ since the performance saturates around this value. The fusion rate $\alpha$ controls the weight of MLLM's prediction in the proposed $\mu$-distance. From Figure~\ref{fig:vis_param_analysis} (b), we find that a mild fusion with the MLLM-based distance benefits the performance most. To get the best result, we choose $\alpha=0.2$. The temperature $\tau$ controls the confidence of the MLLM prediction. As shown in Figure~\ref{fig:vis_param_analysis} (c), we discover that a larger $\tau$ tends to bring a better performance, but a too large value causes degeneration. We choose $\tau=5$ for balancing mAP and Rank-1 performances.

\subsection{Discussion on Computational Cost}
\label{computational_cost}

\begin{figure*}[tbp]
	\centering
	\includegraphics[width=\textwidth]{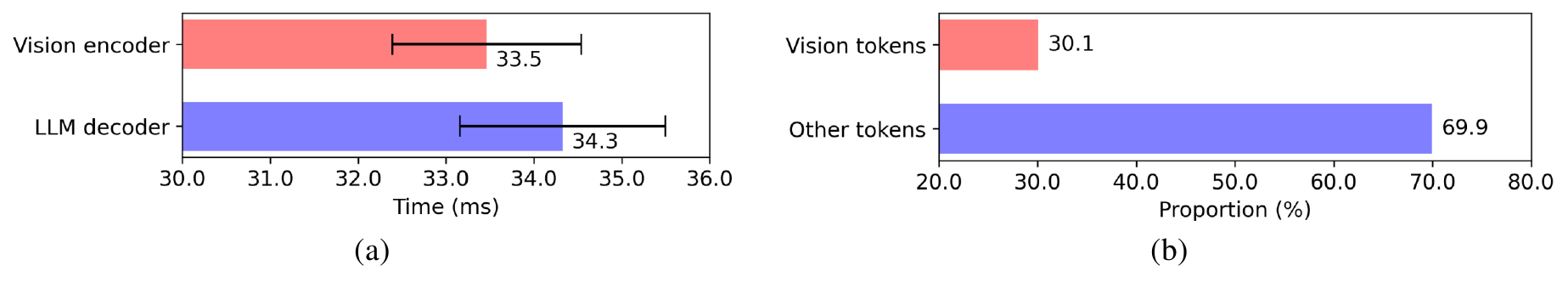}
	\caption{Comparisons on time consumption and input token constitution. (a) illustrates the average inference time of different MLLM components, where the standard deviations are marked out with black bars. (b) illustrates the relative token proportion of different token types in the input sequence.}
	\label{fig:computational_cost}
\end{figure*}

\begin{figure*}[tbp]
	\centering
	\includegraphics[width=0.9\textwidth]{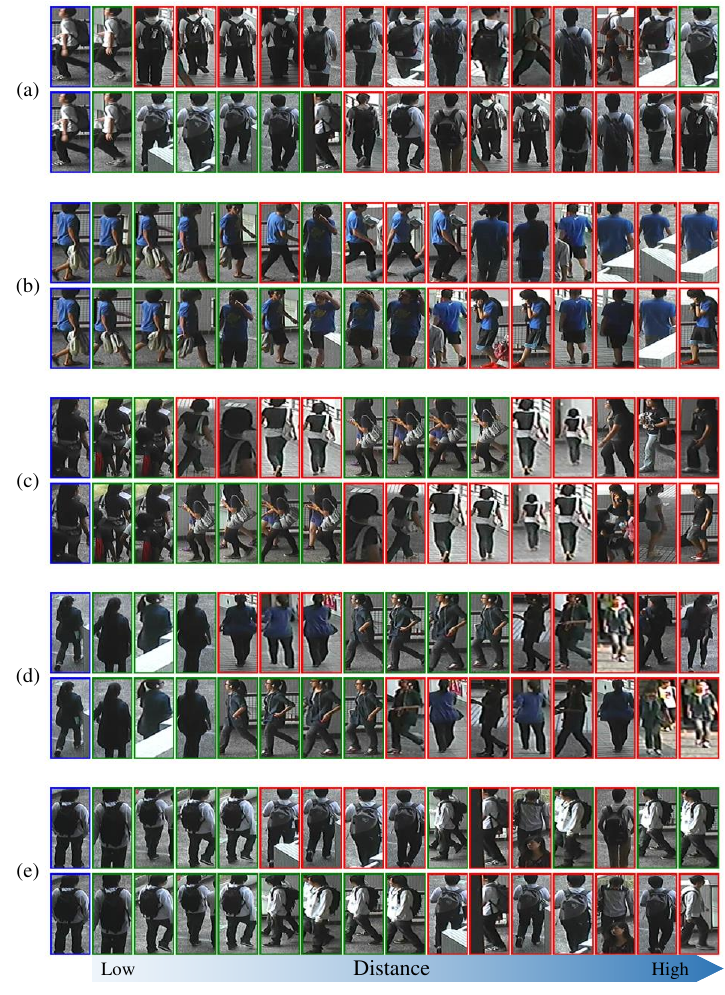}
	\caption{Visualization of ranking lists. Blue, green, and red boxes indicate query image, true matching, and false matching, repectively. Distances are ascending from left to right. In each sample, the first row contains results before enhancement and the second row contains results after enhancement.}
	\label{fig:vis_rank_list}
\end{figure*}

Although our MLLM-based approach enhances the performance of re-ranking methods for DG Re-ID effectively, just like two sides of a coin, it inevitably introduces additional computational overhead. Given an input containing a pair of query-candidate images and a domain-agnostic prompt, we investigate the time consumption of different MLLM components statistically as presented in Figure~\ref{fig:computational_cost} (a) using a single GPU with a batch size of 1.

A complete inference takes 67.8ms in average. We find that the LLM decoder takes more time than the vision encoder in the MLLM. In this work, we adopt parallel inference to reduce the time consumption. In our experiments, the MLLM infers on 4 GPUs parallelly with a batch size of 2 per device, taking about 339.0ms to handle $C=40$ candidates per query with a RAM usage of 5689MB per GPU, which is equivalent to handling an input in about 8.5ms in average.

Moreover, we claim that the inference speed can be further accelerated on a single GPU by caching before inference. The pool of all gallery images can be cached before inference, where the time of candidate image tokenization through the vision encoder can be saved.  Besides,  a large proportion of input tokens are non-vision tokens as illustrated in Figure~\ref{fig:computational_cost} (b). These tokens are fixed and only the vision tokens are altering when inputting different image pairs. It will significantly save time if their corresponding lattent embeddings can be cached in advance for faster LLM decoding.

\subsection{Visualization Results}

In Figure~\ref{fig:vis_rank_list}, we visualize top-15 retrievals of a query image sampled from the target domain testing set of MS$\to$C3 generalization using K-RNN~\cite{CUHK03-NP} re-ranking. Before employing our approach, the re-ranking result still remains suboptimal, where the hard negatives with similar appearances but different IDs are falsely retrieved at front positions. While our approach effectively rectifies them with the MLLM's prediction, where the fine-grained information such as glasses, handbag, and decorations on T-shirt and shoes are focused to make more accurate decisions.

\section{Conclusion}
\label{sec:conclusion}

In this work, we have presented an MLLM-empowered distance metric for re-ranking improvement in DG Re-ID. Specifically, we first adapt an MLLM to the DG Re-ID task by supervised fine-tuning with domain-agnostic prompt and query-candidate hard mining. During inference, the adapted MLLM is employed to yield a robust $\mu$-distance against domain shifts by fusing the raw Euclidean distance and the MLLM-based distance, which can be seamlessly integrated into multiple re-ranking methods and improve their performances significantly. Extensive experiments are conducted to validate the effectiveness of our approach.


{
    \small
    \bibliographystyle{ieeenat_fullname}
    \bibliography{main}
}

\appendix

\section{Performance on Target Domain MA}
\label{appendix:perf_tgt_MA}

In Section~\ref{sec:perf_dg}, the training set of domain MA is used to adapt the MLLM with supervised fine-tuning. To avoid possible domain-leakage problem, where the domain for MLLM adaptation is used as the target domain for testing in DG Re-ID, in the main paper we do not report the results on target domain MA. As you may concern about it, we supplement such results in Table~\ref{tab:perf_tgt_MA_ssdg} and Table~\ref{tab:perf_tgt_MA_msdg} for single-source and multi-source generalization, respectively.

\begin{table}[hb]
	\centering
	\caption{Re-ranking performances on target domain MA with single-source generalization protocol. The MLLM is also fine-tuned on MA. Best results are emphasized in bold and second-best results are highlighted with underline.}
	\label{tab:perf_tgt_MA_ssdg}
	\resizebox{\linewidth}{!}{
		\begin{tabular}{lccccc}
			\toprule
			\multirow{2}{*}{Model} & \multirow{2}{*}{Re-ranking} & \multicolumn{2}{c}{MS$\to$MA} & \multicolumn{2}{c}{C3$\to$MA} \\ \cmidrule{3-6} 
			&  & mAP & Rank-1 & mAP & Rank-1 \\ \midrule
			Baseline~$_{\textit{Ours}}$ & \XSolidBrush & 50.9 & 76.8 & 51.8 & 75.3 \\
			Baseline + K-RNN~\cite{CUHK03-NP}~$_{\textit{CVPR'17}}$ & \Checkmark & 63.0 & 77.1 & 66.7 & 78.7 \\
			Baseline + ECN~\cite{ECN}~$_{\textit{CVPR'18}}$ & \Checkmark & 68.4 & 78.9 & 71.1 & 80.5 \\
			Baseline + CAJ~\cite{CAJ}~$_{\textit{CVPR'24}}$ & \Checkmark & 68.1 & 79.3 & 74.4 & 82.1 \\ \midrule
			Baseline + MUSE~$_{\textit{Ours}}$ & \XSolidBrush & 66.9 & \bf 87.4 & 64.7 & 87.4 \\
			Baseline + MUSE + K-RNN~$_{\textit{Ours}}$ & \Checkmark & 76.9 & 85.2 & 80.7 & 87.8 \\
			Baseline + MUSE + ECN~$_{\textit{Ours}}$ & \Checkmark & \underline{80.1} & 86.6 & \underline{82.5} & \underline{88.7} \\
			Baseline + MUSE + CAJ~$_{\textit{Ours}}$ & \Checkmark & \bf 80.8 & \underline{87.0} & \bf 83.9 & \bf 89.7\\
			\bottomrule
	\end{tabular}}
\end{table}

\begin{table}[hb]
	\centering
	\caption{Re-ranking performances on target domain MA with multi-source generalization protocol. The MLLM is also fine-tuned on MA. Best results are emphasized in bold and second-best results are highlighted with underline.}
	\label{tab:perf_tgt_MA_msdg}
	\resizebox{\linewidth}{!}{
		\begin{tabular}{lccc}
			\toprule
			\multirow{2}{*}{Model} & \multirow{2}{*}{Re-ranking} & \multicolumn{2}{c}{MS+C3+CS$\to$MA} \\ \cmidrule{3-4} 
			&  & mAP & Rank-1 \\ \midrule
			Baseline~$_{\textit{Ours}}$ & \XSolidBrush & 62.3 & 81.4 \\
			Baseline + K-RNN~\cite{CUHK03-NP}~$_{\textit{CVPR'17}}$ & \Checkmark & 74.3 & 82.8 \\
			Baseline + ECN~\cite{ECN}~$_{\textit{CVPR'18}}$ & \Checkmark & 78.9 & 84.7 \\
			Baseline + CAJ~\cite{CAJ}~$_{\textit{CVPR'24}}$ & \Checkmark & 81.9 & 86.7 \\ \midrule
			Baseline + MUSE~$_{\textit{Ours}}$ & \XSolidBrush & 70.8 & 89.6 \\
			Baseline + MUSE + K-RNN~$_{\textit{Ours}}$ & \Checkmark & 83.1 & 88.7 \\
			Baseline + MUSE + ECN~$_{\textit{Ours}}$ & \Checkmark & \underline{86.2} & \underline{90.5} \\
			Baseline + MUSE + CAJ~$_{\textit{Ours}}$ & \Checkmark & \bf 87.7 & \bf 91.4 \\
			\bottomrule
	\end{tabular}}
\end{table}


\end{document}